\documentclass[conference]{IEEEtran}
\IEEEoverridecommandlockouts

\RequirePackage[absolute]{textpos}
\DeclareRobustCommand*{\copyrightnote}{%
  \begin{textblock}{85}(17.5,256.75)
      \scriptsize{\noindent \copyright 2021 IEEE. Personal use of this material is permitted. Permission from IEEE must be obtained for all other uses, in any current or future media, including reprinting/republishing this material for advertising or promotional purposes, creating new collective works, for resale or redistribution to servers or lists, or reuse of any copyrighted
component of this work in other works.}%
  \end{textblock}%
    }

\usepackage{amsmath,amssymb,amsfonts}
\usepackage{algorithmic}
\usepackage{graphicx}
\usepackage[hidelinks]{hyperref}
\usepackage{textcomp}
\def\BibTeX{{\rm B\kern-.05em{\sc i\kern-.025em b}\kern-.08em
    T\kern-.1667em\lower.7ex\hbox{E}\kern-.125emX}}
    
\usepackage[dvipsnames]{xcolor}
\definecolor{vsusp}{HTML}{D62828}
\definecolor{susp}{HTML}{FF931F}
\definecolor{unexp}{HTML}{4169E1}
\definecolor{exp}{HTML}{008080}

\newcommand{\scoreObserved}{Observed}
\newcommand{\scoreTwo}{{\color{exp}Expected}}
\newcommand{\scoreOneTwo}{{\color{unexp}Unexpected}}
\newcommand{\scoreOne}{{\color{susp}Suspicious}}
\newcommand{\scoreZero}{{\color{vsusp}Highly Suspicious}}

\usepackage{placeins}

\usepackage{changes}
\usepackage[nameinlink, capitalize]{cleveref}
\usepackage{todonotes}

\usepackage[giveninits=true,sorting=none,style=ieee]{biblatex}
\AtEveryBibitem{\clearfield{pages}} 

\usepackage{amsfonts, amssymb, amscd}
\usepackage{bbm}
\usepackage{mathtools}
\usepackage{placeins}  
\usepackage[capitalize, nameinlink]{cleveref}
\usepackage{braket}
\newcommand{\T}{\mathrm{T}}
\newcommand{\pderiv}[1]{\frac{\partial}{\partial #1}}

\newcommand{\ent}[1]{\pmb{e}_{#1}}
\newcommand{\stime}[1]{\pmb{t}_{#1}}

\newcommand{\entT}[1]{\pmb{e}^\T_{#1}}
\newcommand{\rel}[1]{\pmb{R}_{#1}}
\newcommand{\vrel}[1]{\pmb{r}_{#1}}
\newcommand{\score}[3]{\theta_{#1,#2,#3}}

\newcommand{\sigmascore}[3]{\sigma \left(\score{#1}{#2}{#3} \right)}
\newcommand{\sigmaeta}[1]{\sigma_\eta\left(#1\right)} 
\newcommand{\sigmaetaform}[1]{\mathrm{min}\left(1, \frac{1}{\eta^2 + e^{-#1}}\right)}

\newcommand{\thresh}{u_\mathrm{th}}
\newcommand{\nlifsum}[1]{\sum_{t_{#1} \leq t^*} W_{s,i#1}}
\newcommand{\nlifsumi}[1]{\sum_{t_{#1} \leq t_{s,i}} W_{s,i#1}}

\newcommand{\triple}[3]{($#1$,\ \nolinebreak $#2$,\ \nolinebreak $#3$)}

\newcommand{\citep}[1]{\cite{#1}}
\newcommand{\citet}[1]{\cite{#1}}

\usepackage{nameref}

\begin{document}

\title{An energy-based model for neuro-symbolic reasoning on knowledge graphs}

\author{\IEEEauthorblockN{Dominik Dold$^{\dagger}$,\ Josep Soler Garrido$^{\dagger}$}
\IEEEauthorblockA{\textit{Siemens AI Lab, Siemens AG Technology}, Munich, Germany
}\vspace{-6.85mm}
\thanks{$^\dagger$ When this research was conducted both authors were with Siemens AG, Munich, Germany. DD is now with the European Space Agency, Noordwijk, Netherlands and JSG is with the European Commission, Joint Research Centre (JRC), Seville, Spain.}
}

\maketitle
\copyrightnote
\thispagestyle{plain}
\pagestyle{plain}
\begin{abstract}
Machine learning on graph-structured data has recently become a major topic in industry and research, finding many exciting applications such as recommender systems and automated theorem proving.
We propose an energy-based graph embedding algorithm to characterize industrial automation systems, integrating knowledge from different domains like industrial automation, communications and cybersecurity.
By combining knowledge from multiple domains, the learned model is capable of making context-aware predictions regarding novel system events and can be used to evaluate the severity of anomalies that might be indicative of, e.g., cybersecurity breaches.
The presented model is mappable to a biologically-inspired neural architecture, serving as a first bridge between graph embedding methods and neuromorphic computing -- uncovering a promising edge application for this upcoming technology.

\end{abstract}

\section{Introduction}\label{sec:introduction}
Multi-relational knowledge graphs (KGs) \citep{singhal2012introducing} are rich data structures used to model a variety of systems like industrial projects \citep{hildebrandt2018configuration} and mathematical proofs \citep{paliwal2020graph}.
It is therefore not surprising that the interest in machine learning algorithms capable of dealing with graph-structured data has increased lately \citep{hamilton2017representation}.
This broad applicability of graphs becomes apparent when summarizing them as lists of triple statements \triple{\text{node}}{\text{edge}}{\text{node}}, e.g., \triple{\text{M.Hamill}}{\text{plays}}{\text{L.Skywalker}} and \triple{\text{L.Skywalker}}{\text{appearsIn}}{\text{StarWars}} -- with individual entries being called subject, predicate and object.
This way, complex relationships between different entities and concepts can be modeled, and machine learning can be used to discover novel facts, i.e., predict or evaluate new edges in the graph.
A widely adapted approach of making the symbolic elements of graphs accessible to machine learning methods are graph embedding algorithms \citep{hamilton2017representation,nickel2015review,ruffinelli2019you}, where nodes and edges of a graph are mapped into a low-dimensional vector space while conserving graph properties like node proximity or link existence.

Graph-based data processing is commonly found in industrial settings where vast amounts of symbolic data from different data silos are combined, stored on servers and used to train models on the cloud \citep{hubauer2018use}.
In contrast, here we are interested in scenarios where graph-structured data is analyzed dynamically, without vast data stores or off-loading to the cloud,
which is generally known as the edge computing paradigm.
One such application can be found in industrial automation systems, where nowadays a convergence of operational technology (OT) and information technology (IT) promises high flexibility, reliability and efficiency in factory systems -- but comes at the price of increasingly complex dynamics between system components and a lack of determinism and isolation \citep{paes2020industrial}.
This leads to novel challenges like guaranteeing system integrity and protecting against cyberattacks, i.e., detecting anomalies in the system before they result in a disruption of operation or even physical damage.

We evaluate a novel graph embedding algorithm for edge computing in such a scenario.
The components of a modern industrial automation system and their complex interaction with each other (\cref{fig:machine}A) can be modeled as a KG \citep{soler2021graph}, taking the form of a list of triple statements with two parts: a static one describing the structure and a dynamic one describing the ongoing interactions between system components (\cref{fig:machine}B).
Data generated this way are incredibly sparse, i.e., only a tiny fraction of possible triples are observed or even valid, as well as streaming in nature such that triples can appear multiple times and underlie stochastic variations.
Using graph embedding, we reformulate the anomaly detection task as a link prediction task: events in the automation system are equivalent to new edges appearing in its graph representation that can be evaluated using the learned embeddings.
However, we found that standard graph embedding algorithms perform poorly on such industrial graphs, mainly because they expect static graphs as input.
In addition, as usually only valid triples are available for training, ``negative samples'' have to be generated from them, introducing possibly erroneous assumptions.
Hence, we introduce a probabilistic, energy-based graph embedding model that is capable of dealing with stochastic variations in graphs and requires no negative samples for training.
The presented model provides sensible likelihood estimations in anomaly detection tasks and can further be mapped to a neuronal representation, opening an exciting prospect for neuromorphic hardware \citep{thakur2018large,davies2018loihi,billaudelle2019versatile,davies2019benchmarks} --  which take inspiration from the brain to achieve energy efficiency and low latency -- as edge learning devices for online graph reasoning.

In the following, we first present the industrial automation demonstrator used to generate application-specific graph data.
Afterwards, we introduce the energy-based graph embedding model, evaluate it on the demonstrator system and provide a conceptual description of how the model could be mapped onto a neuromorphic edge device.
\section{Industrial knowledge graphs}\label{sec:KG}
\begin{figure*}[ht]
    \centering
    \includegraphics[width=0.95\textwidth]{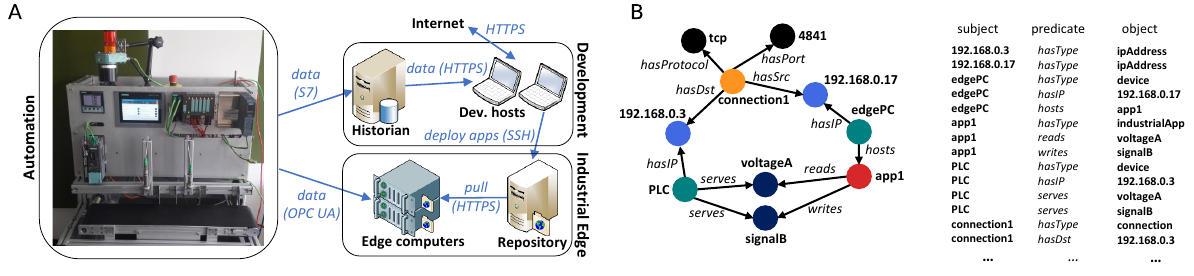}
	\caption{\textbf{(A)} Industrial automation demonstrator consisting of an automation part, a development environment and an industrial edge.
	\textbf{(B)} The demonstrator is represented as a multi-relational graph (left), which is equivalent to a list of triples (right). We only show a small graph snippet here for illustration.
	}\vspace{-4mm}
	\label{fig:machine}
\end{figure*}
We employ an industrial automation demonstrator that mimics the integration of OT and IT technologies in modern industrial systems.
This way, we are able to generate data that captures the complexity and characteristics of such systems while having a controllable and flexible data generation process.
On the OT side, the demonstrator consists of a Siemens S7-1500 Programmable Logic Controller (PLC) connected to multiple subsystems like the drive of a conveyor belt, industrial cameras, a human-machine interface (HMI) and sensors (\cref{fig:machine}A, left).
Information about the system's internal state and sensor measurements are made accessible to edge devices -- dedicated and tightly integrated computing resources -- via an OPC UA server \citep{mahnke2009opc} by the PLC. 
On the IT side, applications (apps) are dynamically hosted on edge computers that access and process this data (\cref{fig:machine}A, right).
An app deployment cycle is realized by a development environment, consisting of several developer hosts that regularly push apps to an app repository based on data made available by a historian -- a database that stores process data from the automation system.
Novel apps are then regularly pulled via HTTPS and hosted on the edge computers.
To increase the degree of realism, the dev. hosts access the internet from time to time.

All three components of the demonstrator system -- engineering data describing the automation part, app activity through the OPC UA server and network events -- can be modeled as multi-relational graphs and consequently combined in a single KG.
Thus, a machine learning algorithm operating on this KG can potentially utilize all three domains to make novel predictions and evaluate unseen activity.
More details on the industrial demonstrator and how to model it as a KG can be found in \citep{soler2021graph}.
\section{Energy-based tensor model}\label{sec:KG}
To analyze the aforementioned data, we propose a probabilistic model of graph embeddings based on an energy function that takes inspiration from the widely used graph embedding algorithm RESCAL \citep{nickel2011three}.

\subsection{Multiplicative graph embedding models}

In RESCAL, a graph is represented as a tensor $X_{s,p,o}$, where entries are $1$ if a triple \triple{s}{p}{o} occurs in the graph and $0$ otherwise.
This way, finding graph embeddings can be rephrased in terms of a tensor factorization problem
\begin{equation}
    X_{s,p,o} \overset{!}{=} \entT{s}\rel{p}\ent{o}\label{eq:rescal} \,,
\end{equation}
with each graph entity $s$ being represented by a $N$-dimensional vector $\ent{s} \in \mathbb{R}^N$ and each relation $p$ by a matrix $\rel{p} \in \mathbb{R}^{N\times N}$.
We denote the sets containing all node and relation embeddings $\mathcal{E}$ and $\mathcal{R}$, respectively.
The problem of finding embeddings is then equivalent to minimizing the reconstruction loss
\begin{equation}
    L_\mathrm{SE} = \sum_{s,p,o} \|X_{s,p,o} - \entT{s}\rel{p}\ent{o} \|^2 \,,
\end{equation}
which can be done via gradient descent.
Usually, we are only aware of valid triples and the validity of all other triples is unknown, i.e., cannot be modeled by setting tensor entries to $0$.
Instead, so-called ``negative samples'' are generated from the training data by randomly exchanging either subject or object entity in a data triple \citep{ruffinelli2019you}.
During training, these negative samples are continuously generated and then presented as ``invalid triples'' with tensor entry $0$.
After learning, novel triples are scored using the obtained embeddings
\begin{equation}
    \score{s}{p}{o} = \entT{s} \rel{p} \ent{o} \,. \label{eq:rscore}
\end{equation}
RESCAL can be further regularized, e.g., by constraining $\rel{p}$ to be diagonal, which is generally known as DistMult \citep{yang2014embedding}.

\subsection{Energy function for graphs}

We assign a tensor $X$ representing a graph the energy
\begin{equation}
    E(X) = -\sum_{s,p,o} X_{s,p,o}\, \score{s}{p}{o} \,.
\end{equation}
From this, we define the probability of observing $X$
\begin{align}
    p(X) = \frac{1}{Z}\, e^{- E(X)} \label{eq:rescal_prob} \,, \ \ \ \ \mathrm{with} \ \ Z = \sum_{X'}  e^{- E(X')} \,,
\end{align}
where we sum over all possible graph realizations $X'$.
Here, the $X_{s,p,o} \in [0, 1]$ are binary random variables indicating whether a triple exists, with the probability depending on the score of the triple.
For instance, a triple \triple{s}{p}{o} with positive score $\score{s}{p}{o}$ is assigned a negative energy and hence a higher probability that $X_{s,p,o} = 1$.
This elevates RESCAL to a probabilistic model by assuming that the observed graph is merely a sample from an underlying probability distribution, i.e., it is a collection of random variables.
Since triples are treated independently here and $Z = \prod_{s,p,o} \left(1+\exp\left(\score{s}{p}{o}\right)\right)$, the probability can be rewritten as
\begin{equation}
    p(X) = \prod_{X_{s',p',o'} = 0} \left(1-\sigmascore{s'}{p'}{o'}\right) \prod_{X_{s,p,o} = 1}\sigmascore{s}{p}{o} \,,
\end{equation}
where $\sigma(\cdot)$ is the logistic function.
Thus, the probability of a single triple \triple{s}{p}{o} appearing is given by $\sigmascore{s}{p}{o}$.

\subsection{Wake-sleep learning}

The model is trained by adjusting node and edge embeddings such that the log-likelihood of observed triples is maximized
\begin{align}
    \Delta \rel{k} &\propto \left\langle\pderiv{\rel{k}} \ln p(\mathcal{B})\right\rangle_{\mathcal{B} \in \mathbb{D}}\hspace{-6mm}\,, \hspace{1mm}
    \Delta \ent{k} \propto \left\langle\pderiv{\ent{k}} \ln p(\mathcal{B})\right\rangle_{\mathcal{B} \in \mathbb{D}} \hspace{-6mm} \,,
\end{align}
where $\mathbb{D}$ is a set of (sub)graphs $\mathcal{B}$ available for learning, e.g., mini-batches or time slices of a data graph $\mathcal{D} = \cup_{\mathcal{B} \in \mathbb{D}}\ \mathcal{B}$. $\langle \cdot \rangle_{y \in Y}$ denotes averaging over the elements $y$ of $Y$. 
The update rules can be rewritten as
\begin{align}
    \Delta \rel{p} &\propto \langle \entT{s} \ent{o} \rangle_{\{s,p,o\}\in \mathcal{B}} - \langle \entT{s} \ent{o} \rangle_{\{s,p,o\}\in \mathcal{S}} \,, \label{eq:dR} \\
    \Delta \ent{k} &\propto \langle \rel{p} \ent{o} \rangle_{\{k,p,o\}\in \mathcal{B}} + \langle \entT{s} \rel{p} \rangle_{\{s,p,k\}\in \mathcal{B}} \nonumber \\ 
               &- \langle \rel{p} \ent{o} \rangle_{\{k,p,o\}\in \mathcal{S}} - \langle \entT{s} \rel{p} \rangle_{\{s,p,k\}\in \mathcal{S}} \label{eq:de} \,,
\end{align}
where $\mathcal{S}$ is a list of triples generated by the model itself, e.g., via sampling (see next section).
Thus, different from graph embedding algorithms like RESCAL, no explicit negative samples are required to train the model.
Relations learn to align with the inner product of subject and object embeddings they occur with, while node embeddings learn to align with the latent representation of their counterpart, e.g., $\ent{s}$ learns to align with the latent representation of the object $\rel{p} \ent{o}$ if \triple{s}{p}{o} is in the data.
Both learning rules consist of two phases, a data-driven and a model-driven phase -- similar to the wake-sleep algorithm used to train, e.g., Boltzmann machines \citep{hinton2002training}.
In contrast to the data-driven phase, during the model-driven phase, the likelihood of model-generated triples $\mathcal{S}$ is reduced.

\subsection{Generating triples using sampling}

To generate triples from the model, we use Markov Chain Monte Carlo (MCMC) sampling -- more precisely, the Metropolis-Hastings algorithm \citep{hastings1970monte} -- with negative sampling as the proposal distribution.
For instance, if the triple \triple{s}{p}{o} is in $\mathcal{B}$, we propose a new sample by randomly replacing either subject, predicate or object, creating a corrupted KG $\tilde{\mathcal{B}}$ containing the new sample instead of \triple{s}{p}{o}.
The sample is accepted with probability $T = \text{min}[1,\ p(\tilde{\mathcal{B}}) / p(\mathcal{B})]$, e.g.,
\begin{align}
   T\big((s,p,o) \rightarrow (s,p,q)\big) = \mathrm{min}\left[1, \exp\left(\pmb{e}_s^\T \pmb{R}_p \pmb{\Delta e}_{qo}\right)\right] \,, \label{eq:transitionprob}
\end{align}
with $\pmb{\Delta e}_{qo}=\pmb{e}_q-\pmb{e}_o$.
The transition probability depends on the distance between the embeddings, i.e., if the embeddings of nodes (or relations) are close to each other, a transition is more likely.
This process is ergodic and can be repeated on the new sample to generate a chain of samples, exploring the neighborhood of the data triple under the model distribution.
\section{Experiments}
\subsection{Graph embedding metrics}\label{sec:metrics}

As a first benchmark, we compare the performance of our proposed probabilistic model with RESCAL for different graph-theoretic metrics.
For this, we employ a recording from the demonstrator system with some network and data activity between demonstrator and edge computers \citep{dold2021spikeembed}, leading to a KG that we randomly split with a ratio of 8/2 into mutually exclusive training and test sets, resulting in 12399 training and 2463 test triples with 3529 entities and 39 relation types.

\begin{figure}[t]
    \centering
    \includegraphics[width=0.9\columnwidth]{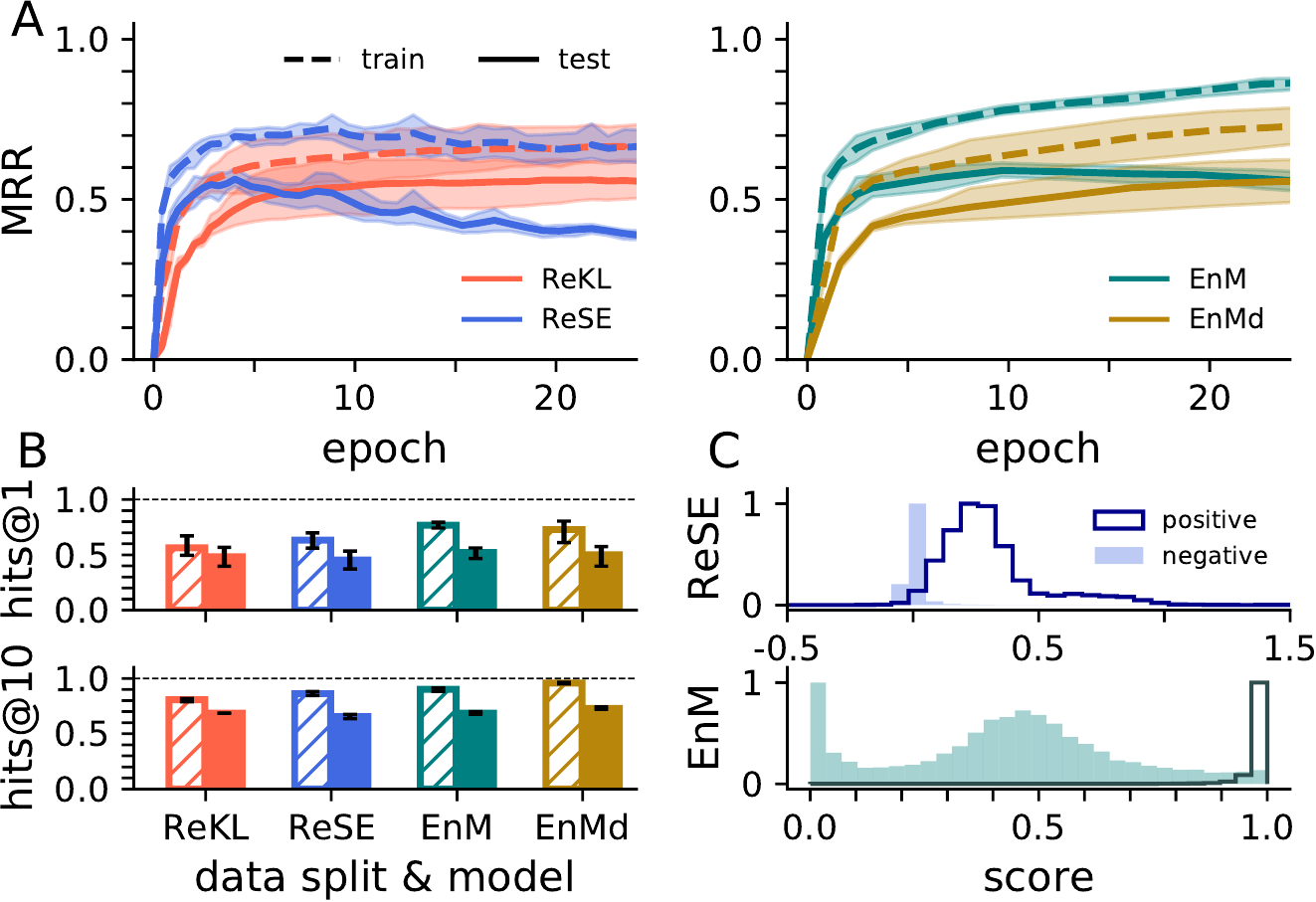}
	\caption{\textbf{(A)} Mean MRR evaluated during training for different models and data splits. 
	Training was repeated for 10 random seeds.
	Shaded areas mark the 15-and 85th percentile.
	\textbf{(B)} Mean hits@k scores.
	Error bars mark the percentiles as in A, and train and test data is represented by striped and filled bars, respectively.
	\textbf{(C)} Distribution of scores $\score{s}{p}{o}$ (top) and probabilities $\sigmascore{s}{p}{o}$ (bottom) for positive and negative validation triples. 
	}\vspace{-4mm}	
	\label{fig:MRR}
\end{figure}

To compare the algorithms, we use the filtered mean reciprocal rank (MRR) and hit-based metrics (hits@k) from the graph embedding literature, see \citep{ruffinelli2019you} for more details.
Both RESCAL (ReSE) as well as our energy-based model without (EnM) and with (EnMd) diagonally-constrained relation matrices achieve similar performance, although our model appears to be less prone to overfitting.
EnMd is especially interesting for neuromorphic implementations, as its symmetric score function enables simple wiring between embedding and output layer (see \cref{fig:network1}A), i.e., no additional mechanism is required to distinguish whether a population acts as subject or object.
We further show that all models are capable of separating positive and negative samples sufficiently (\cref{fig:MRR}C).
In the next section, we introduce RESCAL trained with a Kullback-Leibler loss (ReKL), which is included in \cref{fig:MRR} for completeness.

As an additional comparison, we applied EnM to the well-known UMLS data set -- a biomedical KG encoding the relationships between diseases and chemical compounds -- reaching a total test MRR of $0.80$.
With our implementation of ReSE and ReKL, we reach a similar total test MRR of $0.79$.

\subsection{Context-aware anomaly evaluation}\label{sec:context}
\begin{figure*}[ht]
    \centering
    \includegraphics[width=\textwidth]{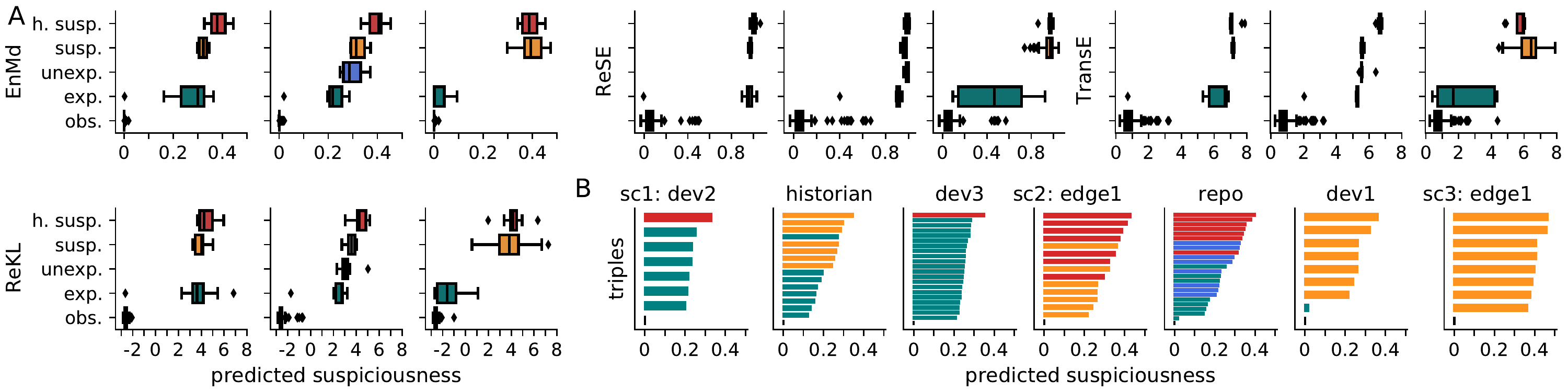}
	\caption{\textbf{(A)} Evaluation of all triples recorded in (from left to right) test scenario (i)-(iii), with class labels given on the y-axis and model predictions on the x-axis. We show the average of 5 simulations with different random seeds. Triples are pooled together according to their class.
	\textbf{(B)} Ranking of triples that only include activity from a specific graph entity, from left to right: HTTPS events by dev. host \#2, with the historian as destination and by dev. host \#3; SSH events involving edge computer \#1, the app repository and by dev. host \#1; network events with edge computer \#1 and app repository as source or destination.
	Triples are sorted here according to EnMd's suspiciousness prediction.
	Colors denote the labelled severity: \scoreZero, \scoreOne, \scoreOneTwo, \scoreTwo\ and Observed. For Observed triples (bottom), i.e., ones that were already seen during the baseline, only the total average is shown.\vspace{-4mm}
	}	
	\label{fig:compare}
\end{figure*}

The primary benchmark of our model is its application in anomaly detection tasks, which we demonstrate here for the industrial automation demonstrator described in \cref{sec:KG}. 
\subsubsection{Experiment setup}
We first record a baseline of normal behavior by running the demonstrator for approximately 50min, resulting in a training graph with 37441 triples, 4347 entities and 38 relation types.
During this time, 5 apps are regularly pulled from the app repository and hosted on 3 edge computers, with apps routinely accessing data from the demonstrator.
At the same time, 3 dev. hosts consistently connect to the historian and the app repository, while also accessing the internet every now and then.

During test time, we alter the behavior of several system components to investigate whether triples generated due to these changes are assigned a decreased likelihood by our model.
This is shown for three separate scenarios here: 
\textbf{(i)} In the first scenario, HTTPS network activity not observed during training is introduced.
These can be events that are expected, e.g., a dev. host might show novel behavior that has been seen for another dev. host during training.
But they might also be suspicious, like a device that is not a dev. host accessing the historian (the historian is usually only accessed by dev. hosts), or even highly suspicious, like an edge device accessing the internet (edge devices should stay in the local network).
\textbf{(ii)} In the second scenario, SSH network activity not observed during training is introduced.
For instance, an expected deviation would be a dev. host accessing the app repository with slightly larger data volume than during training (data volume variations in the local network are usual); SSH activity between edge devices would be highly suspicious (edge devices are never the recipient of SSH connections).
\textbf{(iii)} In the third scenario, edge computer \#2 starts a network scan, leading to many previously unseen connections originating from this device.

In general, we separate system events into five categories depending on their severity: \scoreZero, \scoreOne, \scoreOneTwo, \scoreTwo\ and \scoreObserved, where the severity is based on the behavior defined during the baseline.
A thorough description of the baseline and test cases can be found in \citep{soler2021graph}.

\subsubsection{Evaluation}

The baseline data is integrated into a KG $\mathcal{D}$ and used to train graph embedding models.
The activity of the industrial automation system during test time takes the form of a list of triples, i.e., novel edges appearing in the graph, with 275, 265 and 351 triples for each scenario, respectively.
We use the embeddings ($\mathcal{E}$, $\mathcal{R}$) learned from the baseline ($\mathcal{D}$) to evaluate the severity of these triples, i.e., we estimate how unlikely new edges \triple{s'}{p'}{o'} in the graph are given the structure of the baseline $P(X_{s',p',o'} = 0\ |\ \mathcal{E},\ \mathcal{R}) \approx P(X_{s',p',o'} = 0\ |\ \mathcal{D})$.
For instance, we expect that after training, the node embeddings of dev. hosts will resemble each other (since they share a lot of structure in the baseline KG), and thus activity typical for dev. hosts should be deemed likely by the model, even if it was not observed for one of the dev. hosts during the baseline.

In case of the energy-based model, we define the ``suspiciousness'' of a triple as $S^\mathrm{energy}_{s,p,o} = 1-\sigmascore{s}{p}{o}$, with suspicious activity having high values.
We compare our model with standard RESCAL ($S^\mathrm{SE}_{s,p,o} = 1-\score{s}{p}{o}$) and an alternative probabilistic version of RESCAL trained using a Kullback-Leibler loss $L_\mathrm{KL} = \sum_{s,p,o}  X_{s,p,o}\ \mathrm{log} \frac{X_{s,p,o}}{\mathrm{softmax}\left(\pmb{\theta}\right)_{s,p,o}}$, with $S^\mathrm{KL}_{s,p,o} = -\score{s}{p}{o}$.
As an additional reference, we applied the translating embedding model TransE \citep{bordes2013translating}, where embeddings are found such that $S^\mathrm{TransE}_{s,p,o} = \|\ent{s} + \pmb{r}_p - \ent{o} \| \approx 0$ for valid triples, with relation embeddings $\pmb{r}_p \in \mathbb{R}^N$.

For this benchmark task, we are not looking for a binary anomaly classifier, but rather a severity estimator, i.e., a model that can sort events according to different levels of suspicion.
Typically, such alerts would be sent to a security analyst for further investigation, and thus having meaningful scores is beneficial as it allows prioritization of alerts.
Therefore, we evaluate models by investigating their capability of sorting system events according to their ground truth severity.

\subsubsection{Results}
After training on the baseline, our proposed model is capable of generating well-calibrated estimates in all three test scenarios, with triple likelihoods approximately matching their assigned degree of severity (\cref{fig:compare}A, top left).
This trend appears consistently over a wide range of hyperparameters (not shown here) and is even more pronounced when looking at individual system components (\cref{fig:compare}B):
in this case, we only look at triples that include a certain entity of the automation system as subject or object.
We then rank the triples in a list according to the suspiciousness our model assigns.
As expected, our model is capable of sorting the severity of events correctly, e.g., \scoreZero\ events are on the top, while \scoreTwo\ and \scoreObserved\ ones are on the bottom of the ranked list (\cref{fig:compare}B).

ReSE and TransE perform rather poorly on this task, producing scores for the different suspiciousness levels that are almost binary, i.e., events are either scored high or low (\cref{fig:compare}A, right).
For instance, contrary to EnMd, ReSE and TransE are not capable of separating \scoreTwo, \scoreOne\ and \scoreZero\ events consistently in all scenarios.
This is most likely due to the stochastic nature of our data set, which we test by modeling the data with an alternative probabilistic version of RESCAL using a Kullback-Leibler loss (ReKL), resulting in much better performance (\cref{fig:compare}A, bottom left), but still slightly worse than with the energy-based approach.

The presented results depend on the chosen graph representation of the data, and we are confident that performances can be further improved by choosing representations that are more attuned to the individual benchmark cases.
\subsection{Simulation details}
Simulations were done using Python 3.7.7 and PyTorch 1.6.0.
Embeddings were initialized from a normal distribution $\mathcal{N}\left(\mu, \sigma^2\right)$.
The simulation parameters are given in \cref{tab:exp1}.
Data and code are available on \href{https://www.github.com/dodo47/cyberML}{github.com/dodo47/cyberML}.
\begin{table}[b]\renewcommand{\arraystretch}{1.05}\vspace{-4mm}
\caption{Simulation parameters.}\label{tab:exp1}
\centerline{\resizebox{\columnwidth}{!}{\begin{tabular}{c|cccc|cccc}
\multicolumn{1}{c}{}&\multicolumn{4}{c}{\cref{sec:metrics}}&\multicolumn{4}{c}{\cref{sec:context}}\\
	\hline\hline
	                       & ReSE       & ReKL      & EnM         & EnMd  & ReSE       & ReKL      & EnMd & TransE\\
	\hline
	dimension $N$              & 12         & 12         & 20         & 20   & 20        & 8          & 20  & 8\\
	learning rate $l$          & 0.1        & 0.02       & 0.05       & 0.02 & 0.01        & 0.02       & 0.02  & 0.1\\
	batch size             & 100        & 100        & 200        & 100  & 100        & 100        & 100  & 100\\
	neg. samples $n_\text{s}$           & 2$\times$2 & 2$\times$3 & -          & -   & 2$\times$2  & 2$\times$3 & -  & 2$\times$3\\
	free samples           & -          & -          & 20         & 20   & -           & -         & 20   & - \\
	L1 reg.                & -          & -          & $10^{-4}$  & $10^{-4}$  & -           & -         & -    & - \\
	L2 reg.                & $5\cdot 10^{-5}$ & $10^{-4}$  & -    & -    & $10^{-6}$   & $10^{-6}$ & $10^{-3}$  & $10^{-5}$\\
	$(\mu,\sigma)$     & $(0, 0.1)$   & $(0, 0.1)$         & $(0, 0.1)$ & $(0, 0.1)$     & $(0, 0.1)$   & $(0, 0.1)$          & $(0, 0.1)$ & $(0, 1)$ \\
	optimizer              & Adagrad (AG)    & AG    & AG    & AG & Adam (A) & A & AG  & A  \\
	\hline\hline
	\multicolumn{9}{l}{For TransE, we used a soft margin loss and for ReKL, the softmax norm. was approximated using mini-batches.} \\
	\multicolumn{9}{l}{For UMLS, we used $N=64$ and, in case of ReSE and ReKL, $l = 10^{-3}$, $\text{L2} = 10^{-4}$ and $n_\text{s} = 2\times2$.}
	\end{tabular}}}
\end{table} 
\section{Neuro-inspired representation}
For the suggested application, our algorithm is required to be trained and deployed on the edge to ensure continuous monitoring of industrial automation systems.
An emerging technology for edge computing is neuromorphic hardware \cite{rubino2019ultra,rahimiazghadi2020hardware,covi2021adaptive}, i.e., chips that borrow the architecture of the brain to achieve high energy efficiency and low latency.
However, for an algorithm to be applicable on neuromorphic hardware, it has to take a neuronal form that obeys physical restrictions like locality of information.
In the following, we show that our model can be written in such a form, opening the dynamic world of system monitoring as a potential application area for neuromorphic edge devices.

\begin{figure}[t!]
    \centering
    \includegraphics[width=\columnwidth]{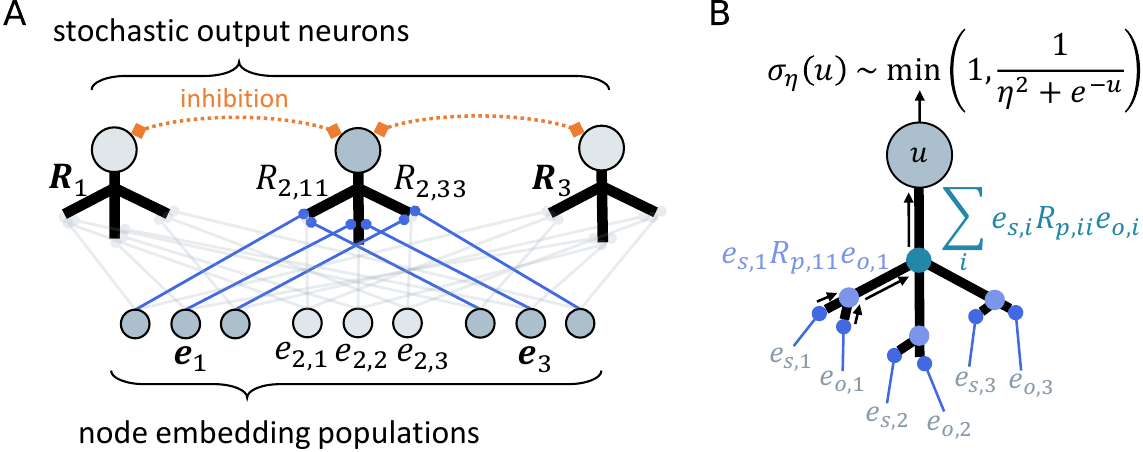}
	\caption{\textbf{(A)} Schematic of the proposed network architecture. 
	Inhibition between the output neurons can be used to ensure that only one triple is returned as output.
	\textbf{(B)} Depiction of the output neurons.
	}	\vspace{-4mm}
	\label{fig:network1}
\end{figure}

The neuronal representation of our model is shown in \cref{fig:network1}A.
Entity embeddings $\ent{x}$ are encoded by populations of neurons, i.e., with one neuron per dimension of $\ent{x}$.
These project statically to output neurons, one for each relation type.
Every output neuron integrates input using a tree-like structure, where each branch encodes a component of the relation embedding $\rel{p}$.
At each of these branches, triple-products of the form $e_{s,i} R_{p,ij} e_{o,j}$ are evaluated and subsequently integrated with contributions from other branches through the tree-like structure (\cref{fig:network1}B).
The integrated input is then fed into an activation function $\sigmaeta{u} = \sigmaetaform{u}$
with $\eta \in [-1, 0, 1]$.
Through $\eta$, the output neurons can return both the probability $\sigma(\cdot)$ of a triple statement to be true ($\eta = -1\ \mathrm{or}\ 1$) and the transition probability $T(\cdot)$ required for sampling ($\eta = 0$).
If both probabilities are evaluated using sampling, the output can be interpreted as stochastic spiking neurons that spike given an input (``triple accepted'') or remain silent (``triple rejected'').

$\eta$ is further used to gate between three different phases for learning: a data and model-driven phase ($\eta = \pm 1$) and a free-running phase ($\eta = 0$) -- which is introduced in the learning rules by adding $\eta$ as a multiplicative factor.
In biologically inspired network models, such a global ``third factor'' \citep{neftci2014event,kusmierz2017learning} is often used to model the effect of neuromodulators, which are known to modulate both the intrinsic firing properties of biological neurons as well as plasticity \citep{marder2012neuromodulation,nadim2014neuromodulation}.
During the free-running phase, the model generates triples which are then replayed to the network in the model-driven phase.
Both during the data and model-driven phase, for each triple \triple{s}{p}{o} parameter updates are calculated, i.e., $\Delta \rel{p} \propto \eta \cdot s_p \cdot \entT{s} \ent{o}$, $\Delta \ent{s} \propto \eta \cdot s_p \cdot \rel{p} \ent{o}$ and $\Delta \ent{o} \propto \eta \cdot s_p \cdot \entT{s} \rel{p}$, 
%
where $s_p$ is the spike response of output neuron $p$, acting as a teaching signal.
For instance during training, if a triple \triple{s}{p}{o} is shown to the network, we excite output neuron $p$ ($s_p = 1$) while inhibiting all other output neurons ($s_{k\neq p}= 0$).
To update the node embeddings, feedback signals are sent from the output neurons to the entity neurons through a pre-wired feedback structure.
Relation embeddings are updated using local information available in the output neurons.
Input is presented to the network by selecting the corresponding embedding populations, which can, e.g., be achieved through gating, resembling a ``memory recall'' of learned concepts that are imprinted into the network through repeated co-activation of embedding populations and output neurons.

A critical property of our model enabling this neuromorphic representation is that predictions are probabilities that can be locally calculated by output neurons.
This is, e.g., not the case for ReSE, where the score is unbounded, and ReKL, where the score of all possible triples is required to calculate probabilities via the softmax operation.
As a downside, embedding neurons are not guaranteed to take strictly positive values and hence are not compatible with a spike-based or even rate-based interpretation -- as required for neuromorphic devices.
Nevertheless, a non-negative variant of RESCAL that might be incorporated in our model has been described in \citep{krompass2013non}.
\section{Conclusion}\label{sec:discussion}
Energy-based models have a long history in neuroscience and artificial intelligence \citep{hopfield1982neural,scellier2017equilibrium}, taking inspiration from physics to describe neuronal dynamics and computations in a structured and compact way.
We introduce an energy-based model for probabilistic modeling of multi-relational graphs.
Different from classical graph embedding algorithms, our model is capable of dealing with the characteristics and stochastic variations of dynamically generated data, like multiple observations of the same triple statement over time.
Furthermore, our model can be trained on true triples only and hence, no assumptions have to be made about how negative samples are generated from true triples in the training data.
In principle, other scoring functions besides the one from RESCAL can be used in the energy function as well.

By rephrasing anomaly evaluation as a link prediction task, we use our proposed model to estimate the severity of anomalous events in an industrial automation demonstrator.
Our model is capable of producing well-calibrated estimates for the severity of anomalous events, and outperforms standard graph algorithms like RESCAL and TransE on this application.
In fact, our model learns to approximate the data distribution, i.e., the typical behavior of the demonstrator, and can consequently be used to detect shifts in the data generation process.
Since triples are human-readable and many system events (like network connections) are represented by multiple triples, we are confident that such estimates can provide security analysts with additional context information and detailed clues to quickly identify and assess system anomalies.

A hallmark of this use case is the dynamic nature and privacy requirements of the recorded graph data that, in practice, require processing on the edge.
Therefore, since our model can be mapped to features of neuromorphic devices, the presented results hint at a novel application area of neuromorphic hardware as edge devices for online graph analytics.

\section*{Acknowledgment}\label{sec:ack}
This work was partially funded by the Federal Ministry for Economic Affairs and Energy of Germany (BMWi) within the IIP-Ecosphere Project.
We thank Marcel Hildebrandt, Serghei Mogoreanu and Martin Ringsquandl for helpful discussions, Johannes Frank for setting up the demonstrator and our colleagues at Siemens SMR and the AI Lab for their support.

\FloatBarrier
\printbibliography
\addcontentsline{toc}{section}{References}

\end{document}